\documentclass{article}

\usepackage[table]{xcolor}
\definecolor{myblue}{RGB}{235, 245, 255}
\usepackage[final]{corl_2026}
\usepackage{times}

\usepackage{booktabs}
\usepackage{multirow}
\usepackage{caption}
\usepackage{colortbl}
\usepackage{amsmath}
\usepackage{amssymb}
\usepackage{amsfonts}
\usepackage{wrapfig} 
\usepackage{threeparttable}

\usepackage{bm} 

\usepackage{graphicx}
\usepackage{algorithm}
\usepackage{algorithmic}
\usepackage{amsmath} 
\usepackage{todonotes}
\usepackage{enumitem}
\usepackage{booktabs}
\usepackage{tabularx}
\usepackage{array}

\usepackage[numbers]{natbib}
\usepackage{multicol}
\setlist{itemsep=2pt, topsep=2pt, parsep=0pt}

\usepackage[capitalize]{cleveref}
\crefname{section}{Sec.}{Secs.}
\Crefname{section}{Section}{Sections}
\Crefname{table}{Table}{Tables}
\crefname{table}{Tab.}{Tabs.}

\newcommand{\mypara}[1]{\vspace{0mm} \noindent \textbf{#1}}
\title{One Agent to Guide Them All: Empowering MLLMs for Vision-and-Language Navigation via Explicit World Representation}

\author{
Zerui Li$^{1,\dagger}$ \quad
Hongpei Zheng$^{2,\dagger}$ \quad
Fangguo Zhao$^{3}$ \quad
Yanyan Shao$^{4}$ \\
Jian Zhou$^{1}$ \quad
Sihao Lin$^{1,6}$ \quad
Shijie Li$^{5}$ \quad
Qi Wu$^{1,6,*}$ \\[6pt]
{\normalfont\normalsize
$^{1}$Australian Institute for Machine Learning, Adelaide University} \\
{\normalfont\normalsize
$^{2}$The University of Manchester \quad
$^{3}$Zhejiang University} \\
{\normalfont\normalsize
$^{4}$Zhejiang Wanli University} \\
{\normalfont\normalsize
$^{5}$Agency for Science, Technology and Research (A*STAR)} \\
{\normalfont\normalsize
$^{6}$Responsible AI Research Centre,} \\
{\normalfont\normalsize
Australian Institute for Machine Learning, Adelaide University} \\[4pt]
{\normalfont\small
$^{\dagger}$Joint first authors \qquad
$^{*}$Corresponding author}
}

\begin{document}
\maketitle


\vspace{-8mm}
\begin{abstract}
A navigation agent must understand high-level semantic instructions and
reason about precise spatial structure. Multimodal Large Language Models
(MLLMs), with their strong generalization, are a promising foundation for
such agents. We propose GTA, a decoupled design that separates low-level
spatial state estimation from high-level semantic planning. We introduce
an explicit metric world representation that combines accumulated geometry
with navigation history and exposes this information to the MLLM through
a multimodal interface for decision-making. We further introduce
counterfactual branch reasoning that checks anticipated outcomes against
subsequent observations and revises navigation decisions when they disagree.
The world representation grounds waypoint selection in reconstructed metric
geometry. GTA achieves 48.8\% Success Rate (SR) on the full R2R-CE val-unseen
split and 46.2\% on a 260-episode English RxR-CE val-unseen subset.
Additional comparisons on shared R2R-CE episodes and mechanism ablations
support the framework's effectiveness. We also demonstrate transfer to
wheeled, quadrupedal, and aerial robots. Project page: \url{https://zeruili22.github.io/GTA/}
\end{abstract}

\section{Introduction}
\vspace{-2mm}
\label{sec:intro}
Vision-and-Language Navigation (VLN)~\cite{anderson2018vision} requires an
embodied agent to follow natural-language instructions in a 3D environment.
Multimodal Large Language Models (MLLMs)~\cite{comanici2025gemini,
bai2025qwen3vltechnicalreport}, with their strong semantic reasoning and
open-world knowledge, are promising generalist VLN
agents~\cite{zhou2024navgpt, zhou2024navgpt2}. Beyond costly fine-tuning that
erodes pretrained knowledge, MLLM-centric \emph{zero-shot} systems are gaining
traction. These systems~\cite{zhou2024navgpt, zhou2024navgpt2, chen2023A2,
qiao2025open} typically adopt a tightly coupled design, jointly performing
spatial and semantic reasoning over long observation streams. However, such designs struggle in challenging scenarios along two dimensions.
At the representation level, relying on egocentric RGB alone forces the
MLLM to infer global spatial structure implicitly, often hallucinating layouts
(e.g., nonexistent room connections). These errors then propagate into semantic
reasoning, yielding instruction-misaligned decisions. At the reasoning
level, processing everything in a single step leaves little room for the
flexible, multi-step strategies needed to fully elicit MLLM capabilities.

\begin{figure}[t]
  \centering
  \includegraphics[width=0.95\linewidth]{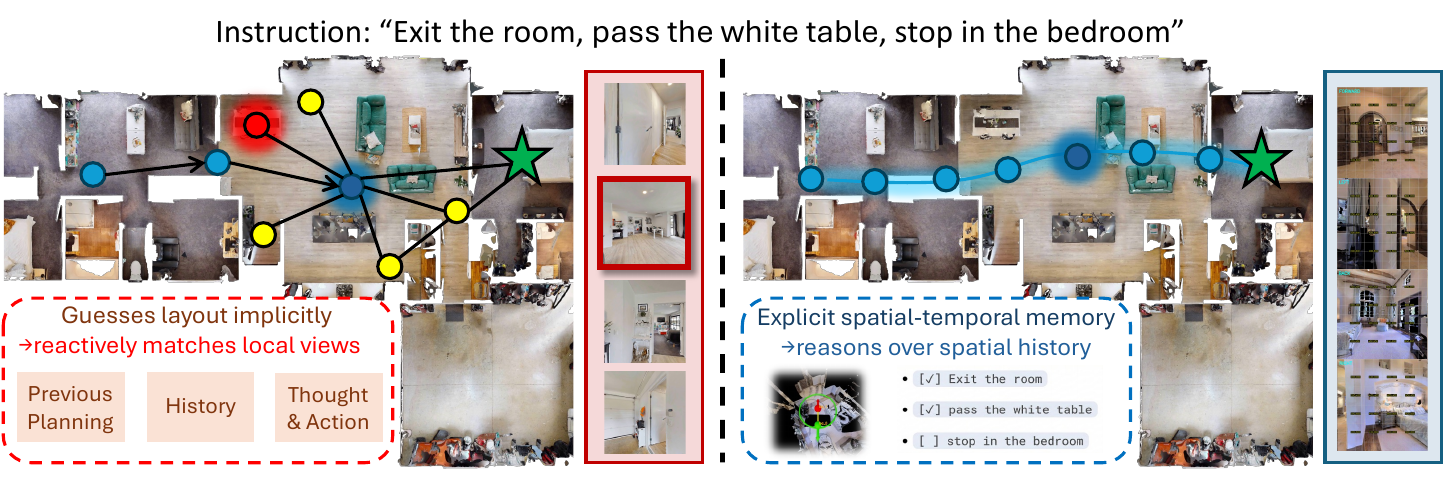}
  \vspace{-4mm}
  \caption{\textbf{Bridging high-level reasoning and embodied execution with GTA.}
  \textit{Left}: Existing MLLM-based VLN agents reduce the rich 3D environment
  into an oversimplified linearized text memory according to human knowledge.
  \textit{Right}: Our GTA framework decouples spatial modeling from semantic
  reasoning and maintains rich spatial and historical information, enabling the
  MLLM to interact with it for decision-making.}
  \label{fig:teaser}
\end{figure}

To address both, we propose GTA (Guide Them All), a zero-shot VLN framework that
decouples low-level spatial state estimation from high-level semantic reasoning
(\cref{fig:teaser}). An explicit metric world representation, kept separate from
the reasoning policy, grounds decisions in a consistent global layout so spatial
errors no longer propagate. This separation in turn lets each part follow its
own reasoning strategy, which a tightly coupled system cannot. Crucially, these
are architectural components. We pair an external, queryable world representation
with a stateful reasoning procedure over it, rather than reformatting
observations into text, and validate their effect by ablations rather than
prompt tuning.
Unlike prior work relying on passive memory buffers or heuristic
maps~\cite{chen2024mapgpt, zhang-etal-2025-mapnav, wang2021structured,
wang2023gridmm}, our world representation lets the agent actively query
task-relevant information on demand, grounding decisions in an explicit layout
and mitigating the hallucinations that plague egocentric reasoning. Building on
this representation, we introduce a counterfactual reasoning strategy that
further elicits MLLM capability. Rather than relying on a fixed waypoint
predictor~\cite{shi2025smartway}, the MLLM evaluates each candidate action by
querying the world representation, reasoning about what each action would entail
against a decomposed goal list, so the chosen action is both globally consistent
and physically valid (\cref{fig:teaser}).

We evaluate GTA on multiple VLN benchmarks~\cite{anderson2018vision,
krantz2020beyond, qi2020reverie}, where it achieves state-of-the-art zero-shot
performance across datasets and MLLM backbones~\cite{bai2025qwen3vltechnicalreport, zhu2025internvl3,openai2025gpt5systemcard, team2023gemini}, and transfers to real robots.
Our contributions are as follows:
\begin{itemize}
    \item We propose GTA, a decoupled zero-shot VLN framework that
    mitigates error propagation and enables task-specific reasoning, achieving
    state-of-the-art zero-shot performance.
    \item We introduce an explicit world representation that gives the
    MLLM global metric context for decision-making, avoiding human-designed
    priors and passive memory summaries.
    \item We develop a counterfactual reasoning strategy operating directly on
    the world representation, enabling physically grounded reasoning that further
    elicits MLLM capability.
\end{itemize}

\vspace{-3mm}
\section{Related Work}
\vspace{-2mm}
\noindent \textbf{Vision-and-Language Navigation (VLN):}
VLN requires an agent to follow natural-language instructions through a visual
environment. Early benchmarks were mostly
discrete~\cite{anderson2018vision, fried2018speaker, qi2020reverie}, reducing
navigation to selecting nodes on a predefined graph. Recent work shifts to
continuous environments (VLN-CE)~\cite{krantz2020beyond} for closer alignment
with real robotics: replacing node ``teleporting'' with low-level actions sharply
degrades traditional models~\cite{krantz2020beyond, irshad_hierarchical_2021,
irshad_sasra_2021, raychaudhuri_language-aligned_2021}, as the agent must now
reason about fine-grained geometry, track its history, and act in a continuous
space. Bridging high-level linguistic reasoning and low-level continuous control
thus remains a core bottleneck.

\noindent \textbf{From Expert Models to Embodied Large Models:}
Early work tackled continuous control with task-specific expert models, notably
waypoint-based architectures~\cite{hong2022bridging, krantz2021waypoint,
li2025ground, wang2023gridmm, an2024etpnav, an2023bevbert}. These improved
success on benchmarks like R2R-CE but act as ``in-domain'' experts tightly
coupled to their training distribution, generalizing poorly across scenes. The
rise of MLLMs shifted the paradigm toward embodied generalists: representative
works adapt LLMs to navigation by mapping visual-linguistic features into the
embedding space and adding navigation-specific tokens, combining in-domain
proficiency with open-vocabulary reasoning~\cite{zhang2024navid, zhang2024uni,
zhang2025embodied}. Yet this adaptation is resource-intensive and, more
critically, structurally couples the agent to a fixed backbone, leaving it unable
to track rapidly improving foundation models~\cite{zhang2025embodied}.

\noindent \textbf{Zero-Shot Embodied Agents in VLN:}
These limitations motivated zero-shot agents that exploit the emergent reasoning
of foundation models without task-specific training. Early efforts showed LLMs
can reason topologically~\cite{chen2024mapgpt, long2024discuss} but stayed in
discrete settings. In VLN-CE, contemporary zero-shot planners represent the
environment by translating observations and topology into serialized text or
language-based mental maps~\cite{chen2023A2, chen2025affordances,
long2025instructnav, chen2025constraint}. Such language-centric representations
preserve a high-level narrative of progress but lack an explicit, geometrically
grounded world model~\cite{qiao2025open, shi2025smartway, li2025boosting}, so
under limited line-of-sight or cluttered geometry the agents struggle to resolve
spatial ambiguity or predict precise waypoints. This gap motivates our explicit
metric world representation. A complementary line studies 3D scene graphs: Hydra~\cite{hughes2022hydra}
constructs hierarchical spatial representations online,
SayPlan~\cite{rana2023sayplan} combines scene-graph search with path planning
and feedback-based replanning, and structured query
interfaces~\cite{ray2026structured} retrieve task-relevant graph information
for language grounding. GTA connects online metric geometry to continuous
instruction following through multimodal prompting, pixel-to-waypoint
grounding, and observation-based verification of branch expectations.

\section{Methodology}
\label{sec:method}

\begin{figure*}[t]
  \centering
  \includegraphics[width=\textwidth]{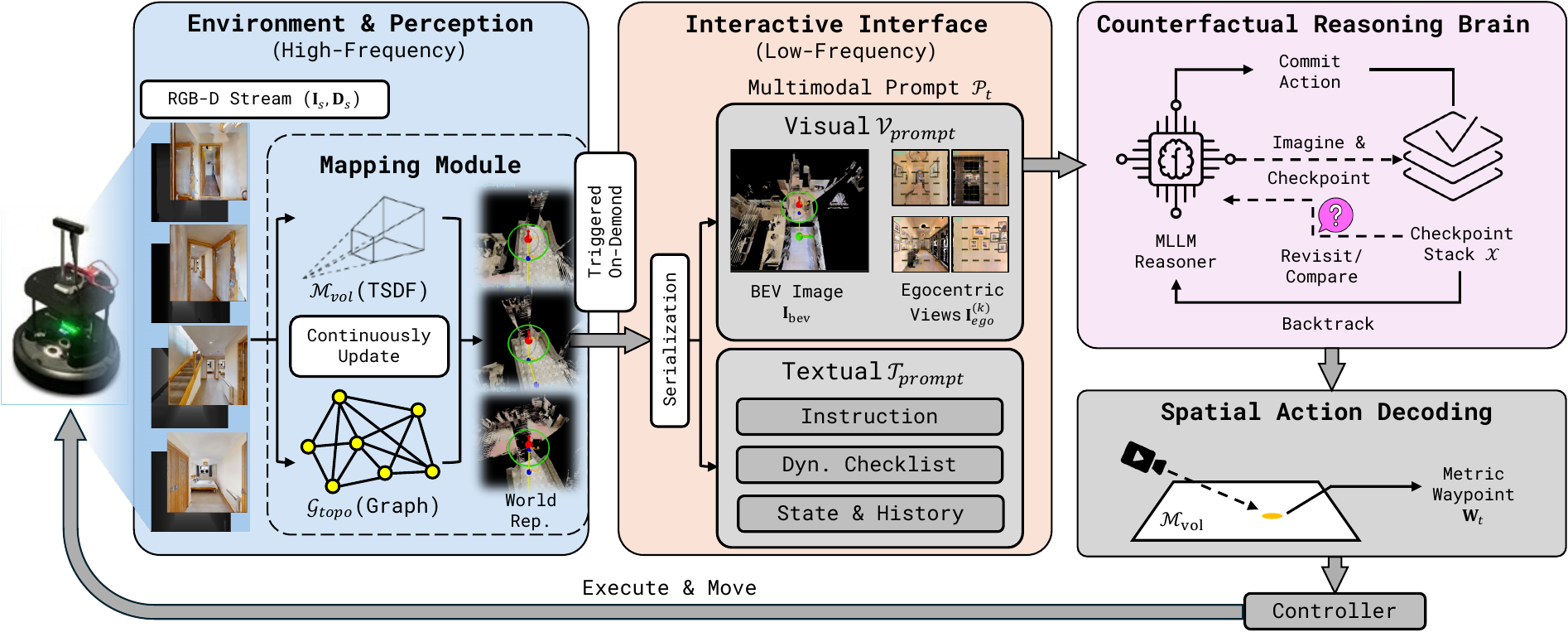}
  \vspace{-2mm}
  \caption{\textbf{Overview of the GTA Framework.} The \textit{Mapping Module} (left) fuses RGB-D streams into a TSDF volume and topological graph, forming our world representation. The \textit{Interactive Interface} (middle) serializes it on demand into a multimodal prompt with BEV/egocentric views, instruction, dynamic checklist, state, and history. The \textit{Counterfactual Reasoning Brain} (right), driven by a frozen MLLM, forms branch-specific expectations, checks them against subsequent observations, and selects actions decoded into metric waypoints.}
  \label{fig:framework}
\end{figure*}

\vspace{-2mm}
We propose GTA, a hierarchical navigation framework that decouples
low-level spatial state estimation from high-level semantic planning
(\cref{fig:framework}). Instead of forcing one model to track geometry and reason
about language at once, GTA maintains an explicit metric world representation
$\mathcal{W}_t$ and lets a frozen MLLM reason over it. At each step $\mathcal{W}_t$
is serialized into a multimodal prompt, the MLLM picks a target, and the target
is decoded into a metric waypoint. Perception runs continuously while reasoning
fires on demand, so decisions rest on an accumulated map rather than a single
noisy frame. We detail the mapping, interface, and reasoning modules in \cref{alg:main}, App.~\ref{app:impl}.

\subsection{Problem Setup}
We address Vision-and-Language Navigation in Continuous Environments (VLN-CE). An
agent placed in a continuous 3D environment receives a natural-language
instruction $\mathcal{I}$, and must reach a target region within a success
threshold from its start pose. Its pose at step $t$ is
$\mathbf{s}_t=(x_t,y_t,\theta_t)$ on the navigable floor plan, with planar
position $\mathbf{p}_t=(x_t,y_t)$ and heading $\theta_t$. The environment is only
partially observed through an asynchronous RGB-D stream. At each reasoning step
the policy emits a structured action $a_t=(m_t,u_t,v_t)$, comprising a view index
$m_t$ and a target pixel $(u_t,v_t)$ within that view, later decoded into a
metric waypoint (\cref{sec:exec}). We assume neither a pre-built map nor a reward
signal. The policy is realized by a frozen, off-the-shelf MLLM $\mathcal{F}$
reasoning on our world representation $\mathcal{W}_t$.

\subsection{Explicit Metric World Representation}
To make the agent aware of both spatial and historical context, the Metric
Mapping Module asynchronously builds a hybrid representation
$\mathcal{W}_t=\langle\mathcal{M}_{vol},\mathcal{G}_{topo}\rangle$, comprising a
volumetric map $\mathcal{M}_{vol}$ that encodes geometry from high-frequency
RGB-D frames and a topological graph $\mathcal{G}_{topo}$ that encodes the
visited trajectory. $\mathcal{W}_t$ accumulates spatial and trajectory context throughout
navigation and provides the spatial input to the interface
(\cref{sec:interface}).

\mypara{Volumetric mapping ($\mathcal{M}_{vol}$).}
To accumulate dense metric geometry across frames, we fuse depth into a TSDF~\cite{curless1996volumetric, newcombe2011kinectfusion}. A
pixel $\mathbf{u}$ at depth $d$ in frame $\mathcal{O}_s$ (yaw $\theta_s$) is
back-projected by $\pi^{-1}(\mathbf{u},d)=d\,\mathbf{K}^{-1}[u,v,1]^{\!\top}$ and
mapped to the world frame via the body pose $\mathbf{T}_{wb}(s)\in SE(3)$ and
extrinsics $\mathbf{T}_{bc}(\theta_s)$, giving world point $\mathbf{p}_w$ and
camera center $\mathbf{p}_{cam}(s)$:
\begin{equation}
    \mathbf{p}_w=\mathbf{T}_{wb}(s)\,\mathbf{T}_{bc}(\theta_s)\,
    \big[\pi^{-1}(\mathbf{u},d);\,1\big],
    \qquad
    \mathbf{p}_{cam}(s)=\mathbf{T}_{wb}(s)\,\mathbf{T}_{bc}(\theta_s)\,[0,0,0,1]^{\!\top}.
\end{equation}
For a voxel $\mathbf{v}$ at $\mathbf{x}$, the truncated projective signed distance is
$\psi_t(\mathbf{x})=\mathrm{clip}\!\big(d-\|\mathbf{x}-\mathbf{p}_{cam}(s)\|,
-\mu,\mu\big)$ with truncation radius $\mu$, fused with a confidence weight
$\omega_t\!\in\![0,1]$ (decaying with depth and grazing angle) by a running average that accumulates into
$\Omega_t(\mathbf{v})$:
\begin{equation}
    S_t(\mathbf{v})=\frac{\Omega_{t-1}(\mathbf{v})\,S_{t-1}(\mathbf{v})+\omega_t\,\psi_t(\mathbf{x})}
                     {\Omega_{t-1}(\mathbf{v})+\omega_t},\qquad
    \Omega_t(\mathbf{v})=\Omega_{t-1}(\mathbf{v})+\omega_t.
    \label{eq:tsdf}
\end{equation}
Its zero-level set defines the reconstructed scene surface, which is reused
for BEV rendering (\cref{sec:interface}) and waypoint ray-casting
(\cref{eq:raycast}).

\mypara{Topological abstraction ($\mathcal{G}_{topo}$).}
The TSDF captures geometry but not where the agent has been. For long-horizon
memory we add a topological graph
$\mathcal{G}_{topo}=(\mathcal{V},\mathcal{E}_{topo})$, where each node
$v_i=\langle\mathbf{p}_i,c_i\rangle$ stores a metric position
$\mathbf{p}_i\in\mathbb{R}^2$ and a visit count $c_i$, and $\mathcal{E}_{topo}$
links consecutive nodes. On reaching $\mathbf{p}_t$, $\textsc{ClusterNode}$
merges the visit into the nearest existing node if within $\delta_{merge}$,
or creates a new node otherwise:
\begin{equation}
    \mathcal{V}\leftarrow
    \begin{cases}
        \text{Update}(v_{nearest}) & d_{min}<\delta_{merge}\\[2pt]
        \mathcal{V}\cup\{\text{Node}(\mathbf{p}_t)\} & \text{otherwise,}
    \end{cases}
\end{equation}
where $d_{min}=\min_{v_i\in\mathcal{V}}\|\mathbf{p}_t-\mathbf{p}_i\|_2$ and
$\delta_{merge}=0.8$\,m. Here $\text{Update}(v_{nearest})$ denotes the node set
with only the nearest node's visit count incremented; its position remains
unchanged. $\text{Node}(\mathbf{p}_t)$ creates a node at $\mathbf{p}_t$ with
initial visit count one. The visit counts $c_i$ make loops and backtracking
explicitly detectable, feeding the safety prompt and counterfactual revision.

\subsection{Interactive Reasoning Interface}
\label{sec:interface}
An MLLM cannot consume $\mathcal{W}_t$ directly, so a training-free
interface $\phi$ serializes it into a multimodal prompt
$\mathcal{P}_t=\{\mathcal{V}_{prompt},\mathcal{T}_{prompt}\}$ with visual and
textual parts. The interface is non-parametric;
\cref{subsec:learnable_interpreter} later replaces it with a learnable
$\phi_\theta$.

\mypara{Egocentric view selection.}
To get $360^\circ$ context without extra actuation, we select four egocentric
views from past frames rather than capturing new ones. From the pool $\mathcal{N}_t$
of the $K{=}20$ spatially nearest past frames, for each target angle
$\alpha_k\in\{0^\circ,90^\circ,180^\circ,270^\circ\}$ we pick the frame whose
orientation best matches $\mathbf{q}_t\!\otimes\!\mathbf{q}(\alpha_k)$ by absolute
quaternion similarity:
\begin{equation}
    s_k=\arg\max_{s\in\mathcal{N}_t}\
    \big|\langle\,\mathbf{q}_s,\ \mathbf{q}_t\!\otimes\!\mathbf{q}(\alpha_k)\,\rangle\big|,
    \qquad
    \big(\mathbf{I}_{ego}^{(k)},\mathbf{D}_{ego}^{(k)}\big)=\big(\mathbf{I}_{s_k},\mathbf{D}_{s_k}\big),
    \label{eq:viewsel}
\end{equation}
each frame assigned to at most one direction, empty if none.

\mypara{Visual prompting with a coordinate grid.}
So the MLLM can emit precise targets without a regression head, we render the
TSDF surface into a bird's-eye view $\mathbf{I}_{bev}$ and overlay an integer
coordinate grid $[0,1000]^2$ on it and the egocentric views, giving
\begin{equation}
    \mathcal{V}_{prompt}=\big[\mathbf{I}_{bev},\ \mathbf{I}_{ego}^{(1)},\dots,\mathbf{I}_{ego}^{(4)}\big].
\end{equation}
The two modalities are complementary. The egocentric RGB views carry appearance
for open-vocabulary grounding of entities named in $\mathcal{I}$, while the BEV
grid carries metric structure for precise waypoint coordinates.

\mypara{Procedural reasoning blueprint ($\mathcal{T}_{prompt}$).}
The textual half is assembled by concatenation ($\oplus$):
\begin{equation}
    \mathcal{T}_{prompt}=\mathcal{C}_t\ \oplus\ \mathrm{State}(\mathcal{G}_{topo},\mathcal{S}_{alert})
    \ \oplus\ \mathrm{History}(\mathcal{H}_t)\ \oplus\ \mathcal{I},
\end{equation}
with the following fields. \textbf{Dynamic checklist $\mathcal{C}_t$:} the
instruction is decomposed into a checklist (e.g.\ ``[x] reach the door'',
``[\,] turn left'') that the MLLM updates each step so the immediate action
stays aligned with the high-level plan. \textbf{Topological \& physical state
$\mathrm{State}(\mathcal{G}_{topo},\mathcal{S}_{alert})$:} the agent's situation in the
graph plus two safety signals collected in $\mathcal{S}_{alert}$, namely vertical
awareness (``upstairs''/``downstairs'' when $|\Delta h|>0.3\,\mathrm{m}$) and
alerts raised when the previous action failed or when a loop is detected
(a node visited at least $c_{\mathrm{loop}}{=}3$ times within a six-step
window, or oscillation between two nodes). \textbf{Execution history
$\mathrm{History}(\mathcal{H}_t)$:} $\mathcal{H}_t$ is the ordered log of recent
(thought, view, action) tuples, truncated to a sliding window of length
$H{=}20$. \textbf{Global instruction $\mathcal{I}$:} the original command, kept
immutable. The assembled $\mathcal{P}_t$ is the sole input to the reasoner in
\cref{sec:brain}.

\begin{figure}[t]
  \centering
  \includegraphics[width=\linewidth]{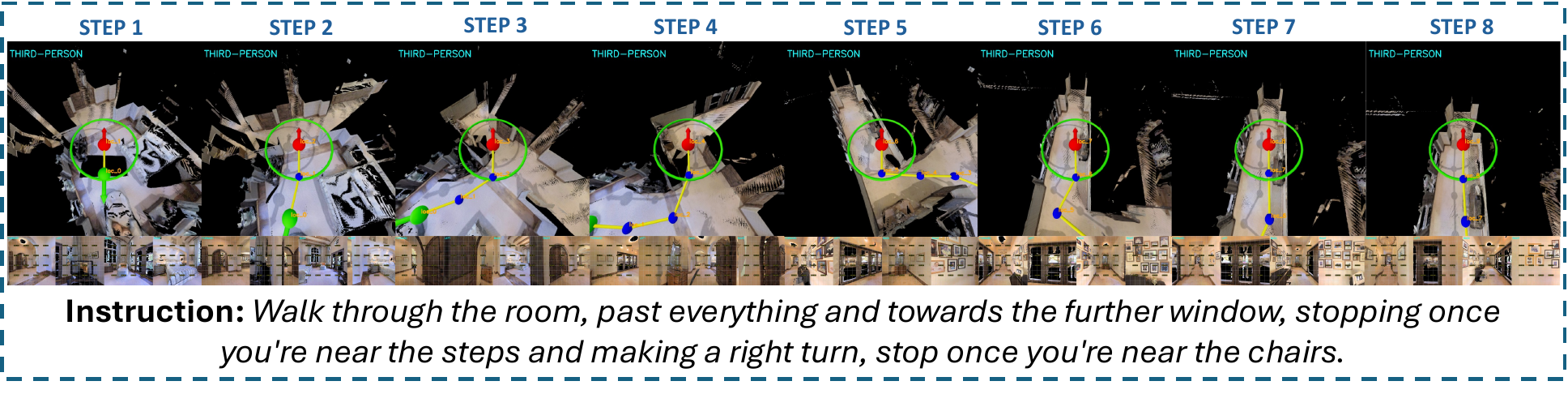}
  \caption{\textbf{Qualitative visualization of a navigation episode in R2R-CE.}
  The top row displays the agent's world representation
  (top-down metric view), while the bottom row shows the four selected
  egocentric views at each step. The planned trajectory is marked in yellow,
  with blue dots indicating waypoints and the red arrow showing the agent's
  current pose.}
  \label{fig:sim_demo}
\vspace{-5pt}
\end{figure}

\subsection{Counterfactual Reasoning Brain}
\label{sec:brain}
Given $\mathcal{P}_t$, the frozen MLLM $\mathcal{F}$ acts as the high-level
policy. A greedy reasoner that commits to one direction per step cannot recover
from an early wrong branch. We therefore let the agent imagine the outcome of
each branch before committing and revisit it when later evidence disagrees.

Specifically, $\mathcal{F}$ uses the egocentric images and rendered map to
identify directions that appear traversable and relevant to the current
checklist item $\mathcal{C}_t$. We denote the resulting viable set by
$\mathcal{B}_t$ and treat the step as a branch point when
$|\mathcal{B}_t|>1$. At such a point, $\mathcal{F}$ emits a per-direction
expectation $e_t^{(k)}$ (e.g., ``a hallway to a kitchen on the right'')
for each $k\in\mathcal{B}_t$ and pushes a checkpoint onto a stack
$\mathcal{K}$,
\begin{equation}
    \kappa_t=\big\langle\,\mathbf{p}_t,\ \{(k,e_t^{(k)})\}_{k\in\mathcal{B}_t},\
    \mathcal{B}_t^{\mathrm{blk}}\,\big\rangle,\qquad \mathcal{B}_t^{\mathrm{blk}}=\emptyset,
\end{equation}
where $\mathcal{B}_t^{\mathrm{blk}}$ records directions already ruled out.
The initially selected branch is $k^\star=m_t$, the view index of the action
emitted at that checkpoint. After advancing along $k^\star$, $\mathcal{F}$
judges, in natural language, whether the observation matches the stored
expectation $e_t^{(k^\star)}$. A rising visit count $c_i$ or a loop alert
from $\mathcal{G}_{topo}$ (\cref{sec:interface}) serves as additional evidence
that the branch is unproductive.

The reasoning becomes counterfactual when the branch is judged inconsistent.
The agent then retrieves the latest active checkpoint $\kappa_{t'}$ and
physically returns to $\mathbf{p}_{t'}$ by retracing visited nodes in
$\mathcal{G}_{topo}$ with the same local planner used for forward motion.
It blocks the failed branch and re-decides over the untried directions,
\begin{equation}
    \mathcal{B}_{t'}^{\mathrm{blk}}\!\leftarrow\!\mathcal{B}_{t'}^{\mathrm{blk}}\cup\{k^\star\},
    \qquad
    k_{t'}=\!\!\arg\max_{k\in\mathcal{B}_{t'}\setminus\mathcal{B}_{t'}^{\mathrm{blk}}}\!
    g\big(k\mid\mathcal{P}_{t'},\mathcal{I}\big),
    \label{eq:counterfactual}
\end{equation}
where $g(\cdot)$ is $\mathcal{F}$'s preference over the remaining directions.
Explicit blocking prevents re-selecting a rejected branch at that checkpoint.
Once a checkpoint's directions are exhausted
($\mathcal{B}_{t'}\setminus\mathcal{B}_{t'}^{\mathrm{blk}}=\emptyset$), it is
released and a safety alert is injected into $\mathcal{T}_{prompt}$ to backtrack
one level further up $\mathcal{K}$ or re-plan; a branch confirmed consistent
retires its checkpoint. This limits repeated exploration at each checkpoint
while allowing recovery from an early wrong branch.

\subsection{Spatial Action Decoding and Execution}
\label{sec:exec}
The MLLM returns an action $a_t=(m_t,u_t,v_t)$, which we turn into a metric goal
by ray-casting from the camera $\mathbf{p}_{cam}^{(m_t)}$ along the viewing ray
$\mathbf{r}^{(m_t)}(u_t,v_t)=\mathbf{K}^{-1}[u_t,v_t,1]^{\!\top}$ (direction only,
distinct from $\pi^{-1}$), transformed to the world frame by the view's
rotation before being cast against the TSDF surface:
\begin{equation}
    \mathbf{w}_t=\mathrm{RayCast}\big(\mathbf{p}_{cam}^{(m_t)},\,\mathbf{r}^{(m_t)}(u_t,v_t),\,\mathcal{M}_{vol}\big),
    \label{eq:raycast}
\end{equation}
thereby grounding the semantic decision in reconstructed metric geometry.
A deterministic local planner attempts to reach $\mathbf{w}_t$; execution
failures and loop alerts are passed to the reasoner on the next step through
the safety mechanism (\cref{sec:interface}).

\subsection{GTA-Light: Extending to a Learnable World Interpreter}
\label{subsec:learnable_interpreter}
\vspace{-1mm}
\mypara{Lightweight neural interpreter.}
The zero-shot system relies on a strong frozen MLLM for high-level reasoning.
Proprietary API-based models (e.g., GPT-5.1~\cite{openai2025gpt5systemcard})
require remote access, while large open-source models
(e.g., Qwen3-VL-235B~\cite{bai2025qwen3vltechnicalreport}) impose substantial
compute requirements. To reduce these deployment costs, we learn a lightweight interpreter
$\phi_\theta$, instantiated on a compact open-source MLLM
(InternVL3-8B~\cite{zhu2025internvl3}), that maps the same world state to the
same prompt space, $\hat{\mathcal{P}}_t=\phi_\theta(\mathcal{W}_t,\mathcal{I})$.
We refer to the full system driven by this distilled interpreter as
GTA-Light. The reasoning protocol is unchanged; only the translator
becomes trainable.

Let $\mathcal{P}_t^\star=(y_1,\dots,y_L)$ be the target prompt from a teacher
MLLM $\mathcal{F}_{tea}$. We fine-tune $\phi_\theta$ with a supervised
sequence-modeling loss over the target tokens,
\vspace{-1mm}
\begin{equation}
    \mathcal{L}_{\text{sft}}(\theta)=
    -\!\!\sum_{(\mathcal{W}_t,\mathcal{P}_t^\star)\in\mathcal{D}}\ \sum_{\ell=1}^{L}
    \log p_\theta\!\big(y_\ell\mid y_{<\ell},\mathcal{W}_t,\mathcal{I}\big),
\end{equation}
and a task-aware decoding loss on the structured action fields (view index
$m_t$, target pixel $(u_t,v_t)$) to sharpen action accuracy,
\begin{equation}
    \mathcal{L}_{\text{act}}=
    \lambda_{\text{view}}\,\mathrm{CE}(\hat{m}_t,m_t^\star)
    +\lambda_{\text{coord}}\big\|(\hat{u}_t,\hat{v}_t)-(u_t^\star,v_t^\star)\big\|_1 ,
\end{equation}
giving $\min_{\theta}\,\mathcal{L}_{\text{sft}}+\mathcal{L}_{\text{act}}$. For
efficiency we use LoRA~\cite{hu2022lora}, freezing $\mathbf{W}_0$ and learning a
low-rank residual $\mathbf{W}=\mathbf{W}_0+\mathbf{B}\mathbf{A}$ with
$\mathrm{rank}(\mathbf{B}\mathbf{A})\ll\min(d_{in},d_{out})$.
QLoRA~\cite{dettmers2023qlora} further back-propagates through a frozen 4-bit
backbone. Training data are generated from the R2R training split by replaying
A$^\star$ trajectories through the world representation and querying the
teacher. No validation scene, episode, or instruction is used for data
construction or student training. We apply input- and prompt-level
robustness augmentations $\mathcal{A}^{W}_{\epsilon},
\mathcal{A}^{P}_{\eta}$ so that the objective becomes
\vspace{-1mm}
\begin{equation}
    \min_{\theta}\ \mathbb{E}_{(\mathcal{W}_t,\mathcal{P}_t^\star)\sim\mathcal{D}}\,
    \mathbb{E}_{\epsilon,\eta}\Big[\mathcal{L}\big(
    \phi_\theta\big(\mathcal{A}^{W}_{\epsilon}(\mathcal{W}_t),\,\mathcal{I}\big),\,
    \mathcal{A}^{P}_{\eta}(\mathcal{P}_t^\star)\big)\Big].
\end{equation}
\vspace{-1mm}
Full data construction and augmentation details are in
Appendix~\ref{app:light}. 
\section{Experiments}
\label{sec:experiments}
\begin{table*}[t]
\centering
\small
\captionsetup{skip=5pt}
\caption{\textbf{Comparison in continuous environments on R2R-CE and RxR-CE Val-Unseen splits.}
The \textbf{best supervised} results are in \textbf{bold}, and the \underline{best zero-shot} results are \underline{underlined}.
We report the full validation split for R2R-CE (1839 episodes) and a sampled subset for RxR-CE (260 episodes).}
\label{tab:continuous_main}
\setlength{\tabcolsep}{9pt}
\resizebox{0.9\linewidth}{!}{%
\begin{tabular}{cl ccccc cccc}
\toprule
\multirow{2}{*}{\textbf{\#}} & \multirow{2}{*}{\textbf{Methods}} & \multicolumn{5}{c}{\textbf{R2R-CE}} & \multicolumn{4}{c}{\textbf{RxR-CE}} \\
\cmidrule(lr){3-7} \cmidrule(lr){8-11} 
& & \textbf{NE}$\downarrow$ & \textbf{OSR}$\uparrow$ & \textbf{SR}$\uparrow$ & \textbf{SPL}$\uparrow$ & \textbf{nDTW}$\uparrow$ & \textbf{NE}$\downarrow$ & \textbf{SR}$\uparrow$ & \textbf{SPL}$\uparrow$ & \textbf{nDTW}$\uparrow$ \\
\midrule
\rowcolor{blue!5} \multicolumn{11}{l}{\textbf{\textit{Supervised Learning:}}} \\
1 & CMA~\cite{hong2021vln} & 6.30 & 49.0 & 38.0 & 33.0 & -- & 10.4 & 24.1 & 19.1 & 37.4 \\
2 & VLN-BERT~\cite{hong2021vln} & 5.74 & 53.0 & 44.0 & 39.0 & -- & 8.98 & 27.1 & 22.7 & 46.7 \\
3 & GridMM~\cite{wang2023gridmm} & 5.11 & 61.0 & 49.0 & 41.0 & -- & -- & -- & -- & -- \\
4 & ETPNav~\cite{an2024etpnav} & 4.71 & 65.0 & 57.0 & 49.0 & -- & 5.64 & 54.8 & 44.9 & 61.9 \\
5 & BEVBert~\cite{an2023bevbert} & 4.70 & 67.0 & 59.0 & 50.0 & -- & 4.80 & 64.4 & -- & 65.4 \\
6 & HNR~\cite{wang2024lookahead} & 4.42 & 67.0 & 61.0 & 51.0 & -- & 5.51 & 56.4 & 46.7 & 63.6 \\
7 & NavFoM~\cite{zhang2025embodied} & 4.61 & 72.1 & 61.7 & 55.3 & -- & 4.74 & 64.4 & \textbf{56.2} & 65.8 \\
8 & Efficient-VLN~\cite{zheng2025efficient} & \textbf{4.18} & \textbf{73.7} & \textbf{64.2} & \textbf{55.9} & -- & \textbf{3.88} & \textbf{67.0} & 54.3 & \textbf{68.4} \\
\midrule
\rowcolor{blue!5} \multicolumn{11}{l}{\textbf{\textit{Zero-Shot Learning:}}} \\
9 & Open-Nav~\cite{qiao2025open} & 6.70 & 23.0 & 19.0 & 16.1 & 45.8 & -- & -- & -- & -- \\
10 & CA-Nav~\cite{chen2025constraint} & 7.58 & 48.0 & 25.3 & 10.8 & -- & 10.37 & 19.0 & 6.0 & -- \\
11 & Smartway~\cite{shi2025smartway} & 7.01 & 51.0 & 29.0 & 22.5 & -- & -- & -- & -- & -- \\
12 & STRIDER~\cite{he2025strider} & 6.91 & 39.0 & 35.0 & 30.3 & 51.8 & 11.19 & 21.2 & 9.6 & 30.1 \\
13 & VLN-Zero$^{\ddagger}$~\cite{bhatt2025vln} & 5.97 & 51.6 & 42.4 & 26.3 & -- & 9.13 & 30.8 & 19.0 & -- \\
14 & BZS-VLN~\cite{li2025boosting} & 6.12 & 55.0 & 41.0 & 25.4 & -- & 7.56 & 35.7 & 21.7 & 42.4 \\
\midrule
\rowcolor{gray!10} 
15 & \textbf{GTA-Light (Ours)}$^\dagger$
& 5.41 & 53.9 & 45.1 & 39.6 & 58.8
& 7.33 & 42.5 & 36.2 & 54.1 \\
\rowcolor{gray!10} 
16 & \textbf{GTA (Ours)}
& \underline{4.95} & \underline{56.2} & \underline{48.8} & \underline{41.8} & \underline{60.4}
& \underline{6.29} & \underline{46.2} & \underline{39.3} & \underline{57.4} \\
\bottomrule
\end{tabular}%
}

\vspace{2pt}
\begin{minipage}{\linewidth}
\footnotesize
$\dagger$~GTA-Light is the lightweight variant (\cref{subsec:learnable_interpreter}), distilled into InternVL3-8B, it is trained and thus excluded from the zero-shot comparison.
\par
$\ddagger$~VLN-Zero pre-explores each scene to construct
its map before evaluation; its results are included
for reference under a different map-access setting.
\end{minipage}

\end{table*}
\vspace{-2mm}

\subsection{Experimental Setup}
\label{subsec:exp_setup}
\vspace{-2mm}
We evaluate on two continuous-environment benchmarks, R2R-CE~\cite{krantz2020beyond}
and RxR-CE~\cite{ku2020room}, in the Habitat simulator~\cite{savva2019habitat},
using the standard VLN-CE metrics (SR, OSR, SPL, NE, nDTW and success within $3$\,m).
Unless specified, the reasoner is GPT-5.1. We further study backbone scalability
in \cref{subsec:ablation}. The main results use the full R2R-CE val-unseen split (1,839 episodes)
and a 260-episode English RxR-CE val-unseen subset. The GTA results in Table~1 are from a representative run
near the median of three runs, with observed spreads of
$\pm0.8$ percentage points in SR and $\pm1.1$ in SPL. The backbone and procedural-blueprint ablations use a curated
180-episode R2R-CE subset. The additional R2R-CE comparisons and
mechanism ablations in \cref{sec:cr-eval} use Open-Nav's released
100-episode set~\cite{qiao2025open, shi2025smartway, he2025strider, zhang2026spatialnav}. We compare against supervised
(\#1--\#8) and zero-shot (\#9--\#14) baselines in \cref{tab:continuous_main}.
Full benchmark, metric, baseline, and subset details are in
Appendix~\ref{app:setup}.

\subsection{Main Results}
\label{subsec:main_results}
\vspace{-2mm}

We present a quantitative comparison of our proposed GTA framework against state-of-the-art baselines in R2R-CE and RxR-CE, as summarized in~\cref{tab:continuous_main}. The results show that world representation substantially improves the navigation performance of generalist MLLMs.

\mypara{State-of-the-Art Zero-Shot Performance}
As evidenced in~\cref{tab:continuous_main}, GTA (\#16) sets a new zero-shot
state-of-the-art on both R2R-CE and RxR-CE, surpassing the prior best zero-shot
agent BZS-VLN~\cite{li2025boosting} (\#14) by 7.8\% SR and 16.4\% SPL on R2R-CE. These results use the same evaluation split but different backbones;
\cref{sec:cr-eval} additionally reports shared-episode comparisons
including STRIDER with GPT-5.1.
The large SPL gain indicates that GTA navigates with map-informed efficiency
rather than exploratory guessing. By decoupling semantic reasoning from spatial
execution and grounding decisions in an explicit metric world representation, the
MLLM produces precise, physically valid waypoints instead of coarse directional
cues. Notably, these gains do not require a frontier backbone. GTA-Light, the same
framework driven by a distilled 8B interpreter
(\cref{subsec:learnable_interpreter}), retains most of the performance (45.1\% SR
/ 39.6\% SPL on R2R-CE, vs.\ 48.8\% / 41.8\%) and still surpasses all prior
zero-shot methods. This indicates that the improvements stem from the framework
itself rather than from the scale of the backbone MLLM.

\mypara{Competitive with Supervised Experts}
Although GTA operates in a strictly zero-shot manner without any parameter updates, it performs comparably to several supervised baselines. As shown
in~\cref{tab:continuous_main}, GTA surpasses classic supervised agents such as CMA~\cite{krantz2020beyond} (\#1) and VLN-BERT~\cite{hong2021vln} (\#2), and
narrows the gap to recent supervised methods like Efficient-VLN~\cite{zheng2025efficient} (\#8). These results suggest that much of the navigation competence needed for VLN already resides in off-the-shelf MLLMs, and that a metric-grounded framework is a promising, complementary direction for eliciting it, rather than a replacement for in-domain training.

\mypara{Robustness on Complex Instructions (RxR-CE).}
RxR-CE features longer trajectories and fine-grained multilingual instructions.
Prior zero-shot agents degrade sharply here. BZS-VLN~\cite{li2025boosting}
(\#14) drops from 41.0\% SR on R2R-CE to 35.7\% on RxR-CE, whereas GTA (\#16)
sustains 46.2\% SR (vs.\ 48.8\% on R2R-CE), and even the distilled GTA-Light
(\#15) holds 42.5\%, still above the prior best. This resilience indicates that
the world representation provides a geometric anchor that buffers
against instructional complexity. An additional 300-episode evaluation is reported in \cref{sec:cr-eval}.
\begin{figure*}[t]
  \vspace{-1em}
  \centering
  \includegraphics[width=\textwidth]{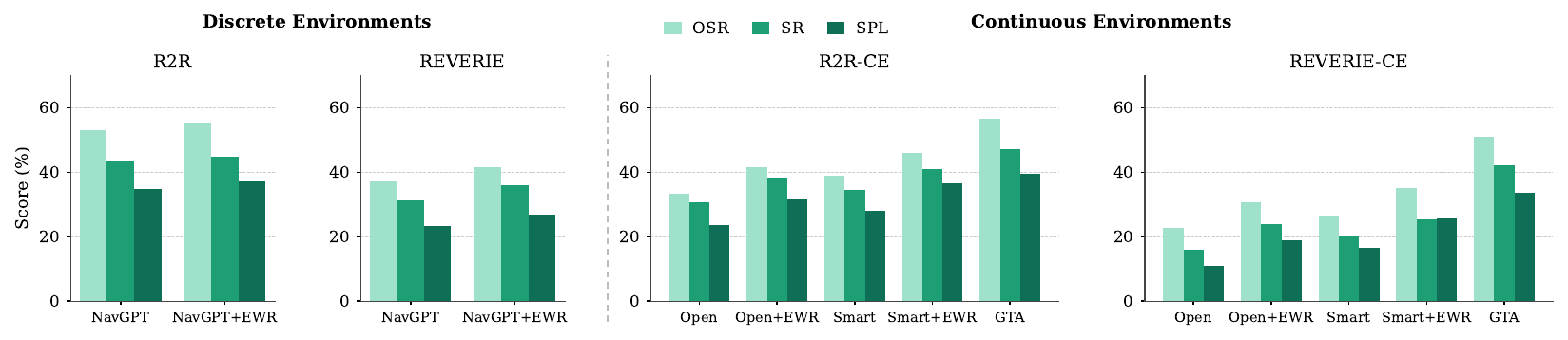}
  \caption{\textbf{Effect of the World Representation across environments.}
  Each panel reports OSR/SR/SPL. Adding WR (world representation) improves every baseline, and GTA leads on all
  metrics in continuous environments. Discrete: R2R / REVERIE; continuous: R2R-CE / REVERIE-CE. Within each continuous-environment comparison, Open-Nav~\cite{qiao2025open}, SmartWay~\cite{shi2025smartway}, and GTA use the same evaluation episodes and the GPT-5.1 reasoner.}
  \label{fig:ewr_ablation}
\end{figure*}

\begin{table}[t]
\vspace{-4mm}
\centering
\small
\begin{minipage}[t]{0.49\linewidth}
\centering
\caption{\textbf{Ablation on MLLM backbones.}}
\label{tab:abl_backbone}
\setlength{\tabcolsep}{3pt}
\resizebox{\linewidth}{!}{%
\begin{tabular}{l|ccccc}
\toprule
\multirow{2}{*}{\textbf{Backbone}} & \multicolumn{5}{c}{\textbf{R2R-CE Val-Sampled} (180)} \\
\cmidrule{2-6}
 & \textbf{NE}$\downarrow$ & \textbf{OSR}$\uparrow$ & \textbf{SR}$\uparrow$ & \textbf{SPL}$\uparrow$ & \textbf{nDTW}$\uparrow$ \\
\midrule
Qwen3-VL-235B  & 5.98 & 46.1 & 37.2 & 29.5 & 51.3 \\
Gemini-2.5 Pro & 5.35 & 48.7 & 42.3 & 34.4 & \textbf{56.2} \\
\rowcolor{gray!10}
GPT-5.1        & \textbf{5.27} & \textbf{56.7} & \textbf{47.2} & \textbf{39.6} & 55.8 \\
\bottomrule
\end{tabular}%
}
\end{minipage}
\hfill
\begin{minipage}[t]{0.49\linewidth}
\centering
\caption{\textbf{Ablation on Procedural Reasoning Blueprints (PB).}}
\label{tab:abl_blueprints}
\setlength{\tabcolsep}{3.5pt}
\resizebox{\linewidth}{!}{%
\begin{tabular}{l|ccccc}
\toprule
\multirow{2}{*}{\textbf{Method}} & \multicolumn{5}{c}{\textbf{R2R-CE Val-Sampled} (180)} \\
\cmidrule{2-6}
 & \textbf{NE}$\downarrow$ & \textbf{OSR}$\uparrow$ & \textbf{SR}$\uparrow$ & \textbf{SPL}$\uparrow$ & \textbf{nDTW}$\uparrow$ \\
\midrule
GTA (w/o PB) & 5.38 & 48.3 & 45.0 & 38.1 & 52.7 \\
\rowcolor{gray!10}
GTA (w/ PB)  & \textbf{5.27} & \textbf{56.7} & \textbf{47.2} & \textbf{39.6} & \textbf{55.8} \\
\bottomrule
\end{tabular}%
}
\end{minipage}
\vspace{-1mm}
\end{table}

\subsection{Ablation Study}
\label{subsec:ablation}

Unless noted otherwise, ablations are conducted on the curated 180-episode R2R-CE subset (\cref{subsec:exp_setup}), so absolute numbers differ from the full-split results in \cref{tab:continuous_main}.

\mypara{Impact of World Representation.}
To verify the necessity of our metric mapping module, we equip baseline agents with and without world representation. As shown in~\cref{fig:ewr_ablation}, world representation yields consistent gains across both discrete and continuous environments, supporting our hypothesis (\cref{sec:intro}) that a structured, metric-accurate description of space helps MLLMs ground semantic decisions rather than rely on reactive visual matching. The
effect is most pronounced on REVERIE and REVERIE-CE, whose coarse-grained object goals (e.g., ``find the pillow in the bedroom'') lack intermediate directional cues. Implicit agents degrade sharply on such tasks for want of a persistent spatial memory, whereas our agents remain robust by reasoning over the mapped structure to infer room connections and unexplored frontiers. This indicates that an explicit spatial representation (world representation) is particularly effective for handling complex, large-scale environments.

\mypara{Scalability with MLLM Reasoning Capabilities.}
By decoupling spatial estimation from decision-making, GTA is model-agnostic and can directly adopt stronger backbones without architectural changes. As shown
in~\cref{tab:abl_backbone}, navigation performance scales with backbone reasoning ability, improving from 37.2\% SR (Qwen3-VL-235B) to 42.3\% (Gemini-2.5 Pro) and
47.2\% (GPT-5.1). This suggests that GTA casts continuous navigation as a high-level reasoning problem, so it stands to benefit as MLLMs improve.

\mypara{Impact of Procedural Reasoning Blueprints.}The w/o PB variant removes exactly the dynamic checklist $\mathcal{C}_t$
and the $\mathrm{State}$ field from $\mathcal{T}_{prompt}$, while retaining
the instruction, history, and visual inputs. As shown in \cref{tab:abl_blueprints},
including PB increases SR from 45.0\% to 47.2\% and OSR from 48.3\%
to 56.7\%, gains of 2.2 and 8.4 percentage points, respectively.
The OSR increase indicates that the agent reaches the target vicinity
more often, even when it does not stop successfully.

\subsection{Additional Comparisons and Mechanism Ablations}
\label{sec:cr-eval}

\begin{table}[t]
\centering

\begin{minipage}[t]{0.48\linewidth}
\vspace{0pt}
\centering
\footnotesize
\captionsetup{skip=5pt}
\setlength{\tabcolsep}{1.5pt}
\caption{Additional comparisons (SR/SPL).
R2R-CE uses the episode sets indicated below.
RxR-CE uses a separate 300-episode sample without
episode-level matching to BZS-VLN.}
\label{tab:cr-compare}
\begin{tabular}{@{}llcc@{}}
\toprule
Protocol & \multicolumn{2}{c}{Baseline} & GTA \\
\midrule
R2R-CE, 100-ep     & STRIDER & 35.0/30.3 & 47.0/38.8 \\
\quad w/ GPT-5.1   & STRIDER & 40.0/36.5 & 47.0/38.8 \\
R2R-CE, 1,839-ep   & BZS-VLN & 41.0/25.4 & 48.8/41.8 \\
RxR-CE, 300-ep     & BZS-VLN & 35.7/21.7 & 46.0/37.9 \\
\bottomrule
\end{tabular}
\end{minipage}\hfill
%
\begin{minipage}[t]{0.50\linewidth}
\vspace{0pt}
\centering
\footnotesize
\captionsetup{skip=5pt}
\setlength{\tabcolsep}{3pt}
\caption{Incremental counterfactual-reasoning ablations on the same
100-episode R2R-CE set; each row adds one mechanism, and (d) is the full GTA.}
\label{tab:cr-cf}
\begin{tabular}{@{}lrrr@{}}
\toprule
Variant & SR$\uparrow$ & SPL$\uparrow$ & nDTW$\uparrow$ \\
\midrule
(a) Greedy, no backtracking   & 31.0 & 27.1 & 34.7 \\
(b) + loop-alert backtrack    & 37.0 & 31.8 & 45.6 \\
(c) + checkpoints, blocking   & 39.0 & 35.3 & 49.3 \\
(d) + expectation checks      & \textbf{47.0} & \textbf{38.8} & \textbf{53.6} \\
\bottomrule
\end{tabular}
\end{minipage}

\end{table}

\textbf{Additional comparisons.}
On Open-Nav's released 100-episode R2R-CE set, GTA achieves
47.0\% SR and 38.8\% SPL (\cref{tab:cr-compare}). Running STRIDER
with GPT-5.1 on the same episodes yields 40.0\% SR and 36.5\% SPL,
leaving gains of 7.0 and 2.3 percentage points.
On an additional random sample of 300 English RxR-CE episodes,
GTA achieves 46.0\% SR and 37.9\% SPL. Since BZS-VLN's episode IDs
were unavailable, its published results provide a reference comparison
without controlling for episode composition.

\textbf{Counterfactual reasoning.}
\Cref{tab:cr-cf} separates three recovery mechanisms.
Variant (a) acts greedily without backtracking. Variant (b) returns
to the last junction on a loop alert but does not remember failed branches.
Variant (c) adds checkpoints and blocked-direction sets, while detecting
failure only through loop or collision signals. Full GTA (d) additionally
checks branch-specific expectations against subsequent observations,
allowing semantic inconsistencies to trigger rejection before a loop alert.
The successive SR gains are 6.0, 2.0, and 8.0 percentage points.
Component ablations are reported in \cref{app:cr-eval}.

\subsection{Real-World Deployment}
\label{subsec:real_world}

\setlength{\intextsep}{2pt}
\begin{wraptable}{rt}{0.36\linewidth}
\centering
\small
\captionsetup{skip=5pt}
\setlength{\tabcolsep}{6pt}
\caption{\textbf{Real-world navigation performance.} \textbf{Bold} = Ours.}
\vspace{-5pt}
\label{tab:real_robot_results}
\begin{tabular}{l cc}
\toprule
\textbf{Method} & \textbf{SR}$\uparrow$ & \textbf{NE}$\downarrow$ \\
\midrule
\rowcolor{blue!5} \multicolumn{3}{l}{\textbf{\textit{Supervised:}}} \\
VLN-BERT~\cite{hong2021vln} & 16.0 & 5.36 \\
RDP~\cite{wang2025rethinking} & 20.0 & 5.45 \\
\midrule
\rowcolor{blue!5} \multicolumn{3}{l}{\textbf{\textit{Zero-Shot:}}} \\
SmartWay~\cite{shi2025smartway} & 32.0 & 4.85 \\
\midrule
\textbf{GTA -- Wheeled} & \textbf{40.0} & \textbf{3.66} \\
\textbf{GTA -- Quadruped} & \textbf{40.0} & \textbf{3.59} \\
\textbf{GTA -- Drone} & \textbf{42.0} & \textbf{3.50} \\
\bottomrule
\end{tabular}
\end{wraptable}
To validate the practical utility of GTA, we deploy it on three physical
platforms, a wheeled robot, a quadruped, and a quadrotor, in unseen real-world
environments (hardware and protocol in Appendix~\ref{app:real}; qualitative
rollouts in Appendix~\cref{fig:real_demo}). As shown
in~\cref{tab:real_robot_results}, the visual domain gap between simulation and
reality cripples end-to-end methods (VLN-BERT 16\%, RDP 20\%), whereas GTA
sustains 40--42\% SR across all three embodiments. We attribute this to our world
representation. Although the MLLM still observes egocentric RGB, its decisions
are anchored on the synthesized top-down metric map and topological graph, whose
structure is domain-invariant (a wall is a wall in both simulation and reality),
keeping the planner effective and stable across bodies despite the appearance
shift. Reasoning latency remains compatible with real-time control under our
asynchronous design (Appendix~\ref{app:latency}).

\vspace{-3mm}
\section{Conclusion and Limitation.}
\vspace{-2mm}

\noindent \textbf{Conclusion. \& Limitation.} We present \textbf{GTA}, a decoupled zero-shot VLN framework that grounds MLLM
reasoning in an explicit and queryable metric world representation. By separating
spatial state estimation from semantic planning and adding counterfactual branch
reasoning, GTA produces physically grounded waypoints without task-specific
fine-tuning. Experiments on R2R-CE and RxR-CE show state-of-the-art zero-shot
performance, while real-world deployment demonstrates that the same
representation supports transfer across robot embodiments. These findings suggest
that a key bottleneck in MLLM-based embodied navigation lies not only in the
reasoning model itself, but also in how the physical world is represented. GTA inherits the assumptions of the proposed world representation. It
relies on reasonably accurate depth, pose, and semantic perception, and currently
focuses on mostly static scenes. Performance may therefore degrade under severe
localization drift, perception failures, or crowded dynamic environments.
Extending the representation with open-vocabulary 3D perception and dynamic
scene modeling is left for future work. The bounded history window also motivates sparse memory retrieval for
longer-horizon applications.

\clearpage



\bibliography{example}  

\clearpage
\appendix
\section{Experimental Setup Details}
\label{app:setup}
\mypara{Benchmarks.}
The main continuous-navigation evaluations use R2R-CE and RxR-CE
in Habitat~\cite{savva2019habitat}. The world-representation comparison in \cref{fig:ewr_ablation}
additionally includes discrete R2R and REVERIE evaluations.
The main R2R-CE result uses the full val-unseen split of 1,839 episodes.
The main RxR-CE result uses a scene-stratified sample of 260 English
val-unseen episodes with seed 1024. The 180-episode R2R-CE ablation set
filters trivial cases and prioritizes long-horizon, multi-room trajectories.
Additional R2R-CE comparisons and mechanism ablations use Open-Nav's
released 100-episode set~\cite{qiao2025open, shi2025smartway, he2025strider, zhang2026spatialnav}. The additional RxR-CE evaluation uses a random
sample of 300 English val-unseen episodes. BZS-VLN's episode IDs were
unavailable, so this comparison does not establish episode-level matching.

\mypara{Metrics.}
We report Success Rate (SR), Oracle Success Rate (OSR), Success weighted by Path
Length (SPL), Navigation Error (NE), and normalized Dynamic Time Warping (nDTW),
the last assessing trajectory fidelity to the ground-truth path. An episode
succeeds if the agent stops within $3.0$\,m of the target.

\mypara{Baselines.}
We compare against two groups (\cref{tab:continuous_main}): supervised agents
(\#1--\#8) trained on in-domain data~\cite{hong2022bridging, wang2023gridmm,
an2024etpnav, an2023bevbert, wang2024lookahead, zhang2025embodied,
zheng2025efficient}, from CMA and VLN-BERT to the recent state-of-the-art
Efficient-VLN, and training-free zero-shot agents (\#9--\#14)~\cite{qiao2025open,
chen2025constraint, shi2025smartway, he2025strider, bhatt2025vln, li2025boosting},
of which BZS-VLN (\#14) is the prior best.

\section{GTA Implementation Details}
\label{app:impl}

\begin{wraptable}[17]{r}{0.39\linewidth}
\centering
\small
\captionsetup{skip=3pt}
\setlength{\tabcolsep}{4pt}
\caption{GTA zero-shot hyperparameters.}
\label{tab:hparams_zs}
\begin{tabular}{ll}
\toprule
\textbf{Component} & \textbf{Setting} \\
\midrule
TSDF voxel size & $0.05$\,m \\
TSDF truncation $\mu$ & $5\times$vox \\
Volume bounds & $[-15,15]^2$ \\
Merge thr.\ $\delta_{merge}$ & $0.8$\,m \\
Loop thr.\ $c_{loop}$ & 3 \\
History window $H$ & $20$ \\
Floor thr.\ $|\Delta h|$ & $0.3$\,m \\
Camera / FOV & $256^2$/$90^\circ$ \\
View angles & $\{0,90,180,270\}^\circ$ \\
Coordinate grid & $[0,1000]^2$ \\
Step (norm/near) & $0.5$/$0.25$\,m \\
Success thr. & $3.0$\,m \\
Max steps & $50$ \\
Reasoner & GPT-5.1 \\
\bottomrule
\end{tabular}
\vspace{-14mm}
\end{wraptable}
The volumetric map $\mathcal{M}_{vol}$ uses a TSDF whose confidence weight
$\omega_t$ decays with depth and grazing angle, with observations fused via the
running-average update (\cref{eq:tsdf}). The topological graph
$\mathcal{G}_{topo}$ merges a new position into the nearest node when their
distance is below $\delta_{merge}$, and otherwise instantiates a new node. A loop
alert is raised when the same node is visited at least three times within a
six-step window, or when the agent oscillates between two nodes.
At each step we select four egocentric views at the canonical angles
$\{0^\circ,90^\circ,180^\circ,270^\circ\}$ from the rotating RGB-D camera, and
overlay an integer coordinate grid over $[0,1000]^2$ on the BEV and egocentric
images. The zero-shot reasoner is GPT-5.1, accessed via its official API. For
ablations, the 180-episode R2R-CE subset (\cref{subsec:exp_setup}) is sampled
from the val-unseen split, favoring episodes with long reference trajectories
and complex, fine-grained instructions. \Cref{tab:hparams_zs} lists the full
hyperparameters of the zero-shot system, and \cref{alg:main} summarizes the
closed-loop interaction.

\section{GTA-Light Training Details}
\label{app:light}

\mypara{Training-data construction.}
Data are generated from the same world representation as the zero-shot
system, using scenes from the R2R training split. No validation scene,
episode, or instruction is used for this data construction or student training.
For each scene, we sample a start $\mathbf{p}_0$ and target
$\mathbf{p}_{goal}$ and compute a reference trajectory by $A^\star$ search
on the navigable graph. Instructions paired with these trajectories are
generated using ScaleVLN~\cite{wang2023scaling}.
The target is used only for offline data generation; the deployed agent
never accesses $\mathbf{p}_{goal}$ and stops when $\mathcal{F}$ emits
a stop action. We replay each trajectory while building $\mathcal{W}_t$
exactly as in the online setting and query the teacher
$\mathcal{P}^{\star}_t=\mathcal{F}_{tea}(\mathrm{Render}(\mathcal{W}_t),\mathcal{I})$
to obtain the supervised pairs
$\mathcal{D}=\{(\mathcal{W}_t,\mathcal{P}^{\star}_t)\}$.
This turns the expensive interpreter into an offline data engine and
amortizes its reasoning into a compact model. Training takes approximately
70 hours on four H100 GPUs. \Cref{tab:hparams_train} summarizes the training configuration.

\mypara{Robustness augmentation.}
To stay stable under the noisy observations encountered in real navigation, we
perturb both the input state and the target prompt during training. The
input-state perturbations $\mathcal{A}^{W}_{\epsilon}$ comprise pose/heading
jitter, dropped graph nodes, masked TSDF cells, depth noise, and varied render
style. The prompt perturbations $\mathcal{A}^{P}_{\eta}$ comprise reordered
non-causal fields, paraphrased alerts, multi-resolution coordinate quantization,
and alternative valid waypoints along the same A$^\star$ corridor. These expose
the model to localization drift, partial observability, and rendering shifts.

\begin{algorithm}[H]
\caption{Overview of the GTA Pipeline.}
\label{alg:main}
\begin{algorithmic}[1]
\REQUIRE Global instruction $\mathcal{I}$, start pose $\mathbf{p}_0$, frozen MLLM $\mathcal{F}$
\REQUIRE Thresholds: $\delta_{merge}$ (node merging), $c_{loop}$ (loop detection), $T_{max}$ (max steps)
\STATE \textbf{Initialize:} $\mathcal{M}_{vol}\!\leftarrow\!\emptyset$, $\mathcal{G}_{topo}\!\leftarrow\!(\{\mathbf{p}_0\},\emptyset)$, $\mathcal{H}\!\leftarrow\!\emptyset$, checkpoint stack $\mathcal{K}\!\leftarrow\!\emptyset$
\STATE \textbf{Initialize checklist:} $\mathcal{C}_t \leftarrow \textsc{Decompose}(\mathcal{I})$
\WHILE{$\mathcal{F}$ has not emitted \textsc{Stop} \textbf{and} $t < T_{max}$}
    \STATE \textcolor{gray}{\textit{// 1. Metric perception (high-frequency)}}
    \STATE Capture rotating view $\mathcal{O}_s=(\mathbf{I}_s,\mathbf{D}_s,\theta_s)$; update $\mathcal{M}_{vol}$ via TSDF (\cref{eq:tsdf})

    \vspace{0.3em}
    \STATE \textcolor{gray}{\textit{// 2. Topological memory \& safety}}
    \STATE $v_{curr}\leftarrow\textsc{ClusterNode}(\mathcal{G}_{topo},\mathbf{p}_t,\delta_{merge})$;\quad $\mathcal{S}_{alert}\leftarrow\emptyset$
    \IF{$c_{v_{curr}} > c_{loop}$}
        \STATE $\mathcal{S}_{alert}\leftarrow$ ``Potential loop detected''
    \ENDIF

    \vspace{0.3em}
    \STATE \textcolor{gray}{\textit{// 3. Verify previous branch against real observation (\cref{eq:counterfactual})}}
    \IF{$\mathcal{K}\neq\emptyset$ \textbf{and} $\mathcal{F}$ judges observation inconsistent with stored $e_{t'}^{(k^\star)}$, \textbf{or} $\mathcal{S}_{alert}\neq\emptyset$}
        \STATE Pop to latest $\kappa_{t'}\in\mathcal{K}$; block failed branch: $\mathcal{B}_{t'}^{\mathrm{blk}}\!\leftarrow\!\mathcal{B}_{t'}^{\mathrm{blk}}\cup\{k^\star\}$
        \STATE Physically retrace visited nodes in $\mathcal{G}_{topo}$ to $\mathbf{p}_{t'}$ using $\pi_{local}$
        \STATE \textbf{if} $\mathcal{B}_{t'}\!\setminus\!\mathcal{B}_{t'}^{\mathrm{blk}}=\emptyset$ \textbf{then} release $\kappa_{t'}$, inject backtrack alert into $\mathcal{S}_{alert}$
    \ENDIF

    \vspace{0.3em}
    \STATE \textcolor{gray}{\textit{// 4. Interface: serialize $\mathcal{W}_t$ into prompt $\mathcal{P}_t$}}
    \STATE $\{\mathbf{I}_{ego}^{(k)}\}_{k=1}^{4}\leftarrow\textsc{SelectViews}(\mathcal{N}_t)$ \quad (\cref{eq:viewsel})
    \STATE $\mathcal{V}_{prompt}\leftarrow[\,\mathbf{I}_{bev},\{\mathbf{I}_{ego}^{(k)}\}_{k=1}^{4}\,]$
    \STATE $\mathcal{T}_{prompt}\leftarrow\mathcal{C}_t\oplus\textsc{State}(\mathcal{G}_{topo},\mathcal{S}_{alert})\oplus\textsc{History}(\mathcal{H})\oplus\mathcal{I}$
    \STATE $\mathcal{P}_t\leftarrow\{\mathcal{V}_{prompt},\mathcal{T}_{prompt}\}$

    \vspace{0.3em}
    \STATE \textcolor{gray}{\textit{// 5. Reasoning: act, and checkpoint at branch points}}
    \STATE $\mathcal{R}\leftarrow\mathcal{F}(\mathcal{P}_t)$;\quad parse $(m_t,u_t,v_t),\,\mathcal{C}_{t+1}\leftarrow\textsc{ParseJson}(\mathcal{R})$
    \STATE $\mathcal{B}_t\leftarrow$ directions marked by $\mathcal{F}$ as traversable and relevant to $\mathcal{C}_t$ (from $\mathcal{R}$)
    \IF{$|\mathcal{B}_t| > 1$}
        \STATE $\mathcal{F}$ emits per-direction expectations $\{e_t^{(k)}\}_{k\in\mathcal{B}_t}$
        \STATE Push $\kappa_t=\langle\mathbf{p}_t,\{(k,e_t^{(k)})\}_{k\in\mathcal{B}_t},\emptyset\rangle$ onto $\mathcal{K}$
    \ENDIF

    \vspace{0.3em}
    \STATE \textcolor{gray}{\textit{// 6. Grounding \& execution}}
    \STATE $\mathbf{w}_t\leftarrow\textsc{RayCast}(\mathbf{p}_{cam}^{(m_t)},\mathbf{r}^{(m_t)}(u_t,v_t),\mathcal{M}_{vol})$ \quad (\cref{eq:raycast})
    \STATE $\mathbf{c}_t\leftarrow\pi_{local}(\mathbf{p}_t,\mathbf{w}_t)$
    \STATE Execute $\mathbf{c}_t$, advance to $\mathbf{p}_{t+1}$, update history $\mathcal{H}$
\ENDWHILE
\end{algorithmic}
\end{algorithm}

\section{Real-World Deployment Details}
\label{app:real}

\mypara{Wheeled robot.}
For ground-based tasks we use a TurtleBot~4 mobile base equipped with an Intel
RealSense D455 RGB-D camera. The camera is mounted on a servo motor to perform
the orthogonal panoramic scanning (4 views) described in \cref{sec:method}.
Low-level motion planning is handled by the ROS~2 Nav2 stack, while MLLM
inference runs on a remote server communicating via standard HTTP requests.

\mypara{Quadruped robot.}
For legged locomotion we use a Unitree Go2 quadruped equipped with an Intel
RealSense RGB-D camera, mounted on a servo to perform the same orthogonal
four-view scanning as the wheeled platform. We deliberately reuse the wheeled
robot's perception and navigation setup unchanged, including the ROS~2 motion
stack and the same remote MLLM-inference interface, so that the only variable
across the two ground platforms is the locomotion dynamics. This isolates
embodiment as the sole factor and tests whether the world representation
remains a valid interface under the irregular, non-planar motion of a legged
body.

\mypara{Aerial robot.}
For aerial operations we employ a custom-built quadrotor integrated with an
Intel RealSense D435 RGB-D camera. The UAV uses a Betaflight-based flight
controller, while high-level command computation and communication run on a
CoolPi~4B onboard computer. Precise state estimation and global localization are
provided by an external motion-capture system.

\begin{table}[!t]
\centering
\small
\caption{GTA-Light training hyperparameters.}
\label{tab:hparams_train}
\begin{tabular}{ll}
\toprule
\textbf{Component} & \textbf{Setting} \\
\midrule
Student backbone & InternVL3-8B \\
Teacher & GPT-5.1 \\
LoRA rank $r$ / $\alpha$ & $16$ / $32$ \\
Trainable modules & LoRA adapters + multimodal projector \\
Quantization & 4-bit NF4 (QLoRA) \\
Learning rate & $2\times10^{-4}$ \\
Batch size & $128$ effective batch size \\
Epochs & $3$ \\
Optimizer & paged AdamW 8-bit \\
Loss weights $\lambda_{view}$ / $\lambda_{coord}$ & $1.0$ / $2.0$ \\
Training trajectories / pairs & $10$k / $180$k \\
\bottomrule
\end{tabular}
\end{table}

\mypara{Protocol.}
We conduct 50 navigation trials with complex language instructions on each
platform in unseen real-world environments. The three embodiments span a wide
range of motion dynamics and camera viewpoints, from the stable planar motion of
the wheeled base, through the irregular gait-induced jitter of the quadruped, to
the free six-DoF motion and elevated viewpoint of the aerial platform, while
sharing the same GTA framework without platform-specific tuning.

\begin{figure}[t]
  \centering
  \includegraphics[width=\linewidth]{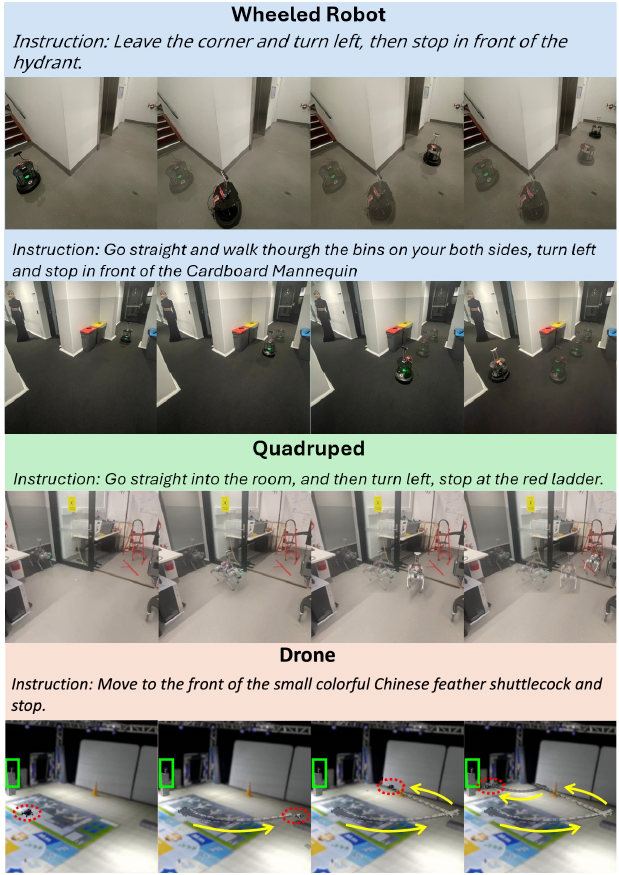}
  \caption{\textbf{Zero-shot sim-to-real transfer across diverse robot
  embodiments.} We demonstrate the generalization of GTA by deploying it on
  distinct physical platforms in unseen real-world environments.}
  \label{fig:real_demo}
\end{figure}

\section{Wall-Clock Latency and Real-Time Viability}
\label{app:latency}

\begin{table}[h]
\centering
\caption{Inference latency per reasoning step, measured over 100 trials.}
\label{tab:latency}
\renewcommand{\arraystretch}{1.1}
\resizebox{0.65\columnwidth}{!}{%
\begin{tabular}{lcc}
\toprule
\textbf{Module / Backbone} & \textbf{Mean Latency (s)} & \textbf{Std. Dev. (s)} \\
\midrule
\rowcolor{blue!5}
\multicolumn{3}{l}{\textit{Perception \& Mapping}} \\
TSDF Integration + BEV Rendering & $0.0972$ & $\pm 0.0844$ \\
\midrule
\rowcolor{blue!5}
\multicolumn{3}{l}{\textit{Reasoning}} \\
GTA-Light (Local)      & $3.96$ & $\pm 0.47$ \\
Gemini-2.5-Flash (API)   & $3.27$ & $\pm 0.37$ \\
GPT-5.1 (API)            & $5.42$ & $\pm 1.92$ \\
\bottomrule
\end{tabular}%
}
\end{table}

\vspace{3mm}
A practical concern for any MLLM-based navigation system is whether the reasoning latency is compatible with real-time deployment. We address this through two complementary arguments: (1) an asynchronous execution paradigm that structurally decouples perception from reasoning, and (2) empirical latency measurements across all backbone configurations.

\mypara{Asynchronous Execution Paradigm.}
GTA adopts a hierarchical, asynchronous design in which the \emph{Perception \& Mapping} module runs at high frequency while the \emph{Counterfactual Reasoning Brain} operates as a low-frequency ``Slow System.'' Rather than blocking navigation on each MLLM call, the system initiates the next reasoning query as soon as the robot departs toward its current waypoint. At a safe indoor cruising speed of $0.1$\,m/s, the robot requires approximately $5$--$10$\,s to traverse a typical inter-waypoint distance of $0.5$--$1.0$\,m. The measured mean reasoning latencies support overlapping inference with motion on typical segments, although latency variability can still cause pauses on shorter segments.

\mypara{Latency Measurements.}
Table~\ref{tab:latency} reports the mean inference latency and standard deviation per reasoning step, measured over 100 trials for each module and backbone. TSDF integration and BEV rendering together take 0.0972~s on average, contributing little overhead. Mean reasoning latency ranges from 3.27~s (Gemini-2.5-Flash) to 5.42~s (GPT-5.1), supporting overlap with motion on typical segments.

\section{Additional Evaluation Details}
\label{app:cr-eval}

\begin{table}[H]
\noindent
\begin{minipage}[t]{0.55\linewidth}
\vspace{0pt}
\normalsize
\textbf{Component ablations.}
On Open-Nav's 100-episode R2R-CE set,
\cref{tab:cr-components} reports SR changes in percentage
points relative to full GTA (47.0\% SR).
Removing egocentric views causes the largest drop,
followed by BEV and topology.
Removing the coordinate grid leaves aggregate SR unchanged
on this set; this does not establish its redundancy
across models or environments.
\end{minipage}\hfill
\begin{minipage}[t]{0.42\linewidth}
\vspace{-6pt}
\centering
\small
\setlength{\abovecaptionskip}{0pt}
\setlength{\belowcaptionskip}{5pt}
\caption{Component ablations}
\label{tab:cr-components}
\begin{tabularx}{\linewidth}{@{}>{\raggedright\arraybackslash}Xr@{}}
\toprule
Ablated component & $\Delta$SR (pp) \\
\midrule
BEV               & $-5.0$  \\
Topological graph & $-3.0$  \\
Egocentric views  & $-16.0$ \\
Coordinate grid   & $0.0$   \\
\bottomrule
\end{tabularx}
\end{minipage}
\end{table}

\section{Notation}
\label{app:notation}

\begin{table}[ht]
\centering
\small
\caption{Main notation used in GTA.}
\label{tab:notation}
\renewcommand{\arraystretch}{1.08}
\begin{tabularx}{\linewidth}{
    @{}l>{\raggedright\arraybackslash}X@{}
}
\toprule
Symbol & Meaning \\
\midrule
$\mathcal{I}$
& Global navigation instruction. \\

$\mathcal{F}$, $\phi$, $\phi_\theta$
& Frozen MLLM reasoner, training-free interface,
and learned interpreter. \\

$\mathbf{s}_t$, $\mathbf{p}_t$
& Agent pose $(x_t,y_t,\theta_t)$ and planar position. \\

$a_t=(m_t,u_t,v_t)$
& Action comprising a view index and target pixel. \\

$\mathcal{W}_t=\langle\mathcal{M}_{vol},\mathcal{G}_{topo}\rangle$
& World representation: TSDF volume and topological graph. \\

$S_t(\mathbf{v})$, $\Omega_t(\mathbf{v})$
& Fused signed distance and accumulated voxel weight. \\

$\psi_t$, $\omega_t$, $\mu$
& Per-observation signed distance, fusion weight,
and truncation radius. \\

$\mathbf{T}_{wb}$, $\mathbf{T}_{bc}$, $\mathbf{K}$
& Body pose, camera extrinsics, and camera intrinsics. \\

$v_i=\langle\mathbf{p}_i,c_i\rangle$
& Topological node storing position and visit count. \\

$\delta_{merge}$, $c_{loop}$
& Node-merging distance and loop-detection visit threshold. \\

$\mathcal{N}_t$, $\mathbf{q}_s$, $\mathbf{q}_t$
& Candidate past frames, frame orientation,
and current agent orientation. \\

$\mathbf{I}_{bev}$, $\mathbf{I}_{ego}^{(k)}$
& Rendered map view and selected egocentric view. \\

$\mathcal{P}_t$, $\mathcal{V}_{prompt}$, $\mathcal{T}_{prompt}$
& Multimodal prompt and its visual and textual components. \\

$\mathcal{C}_t$, $\mathcal{S}_{alert}$, $\mathcal{H}_t$, $H$
& Dynamic checklist, safety alerts, execution history,
and history-window length. \\

$\mathcal{B}_t$, $\mathcal{B}_t^{\mathrm{blk}}$
& Viable branches and blocked directions. \\

$e_t^{(k)}$, $k^\star$, $g(\cdot)$
& Branch expectation, selected branch,
and the MLLM's preference over remaining branches. \\

$\kappa_t$, $\mathcal{K}$
& Checkpoint and checkpoint stack. \\

$\mathbf{w}_t$, $\pi_{local}$
& Grounded metric waypoint and local planner. \\
\bottomrule
\end{tabularx}
\end{table}

\end{document}